\documentclass[conference]{IEEEtran}
\IEEEoverridecommandlockouts

\usepackage{cite}
\usepackage[utf8]{inputenc} 
\usepackage{amsmath,amssymb,amsfonts}
\usepackage{algorithmic}
\usepackage{graphicx}
\usepackage{hyperref}
\usepackage{comment}
\usepackage{textcomp}
\usepackage{xcolor}
\usepackage{url} 
\def\BibTeX{{\rm B\kern-.05em{\sc i\kern-.025em b}\kern-.08em
    T\kern-.1667em\lower.7ex\hbox{E}\kern-.125emX}}
\begin{document}

\newcommand{\fp}[1]{\textcolor{red}{[#1]}}

\title{Puzzled By ChatGPT? No more! \\ {\LARGE A Jigsaw Puzzle to Promote AI Literacy and Awareness}
}

\author{\IEEEauthorblockN{1\textsuperscript{st} Francesca Padovani}
\IEEEauthorblockA{\textit{Center for Language and Cognition (CLCG)} \\
\textit{University of Groningen}\\
Groningen, The Netherlands\\
f.padovani@rug.nl}
\and
\IEEEauthorblockN{2\textsuperscript{nd} Malvina Nissim}
\IEEEauthorblockA{\textit{Center for Language and Cognition (CLCG)} \\
\textit{University of Groningen}\\
Groningen, The Netherlands \\
m.nissim@rug.nl}
}


\maketitle

\begin{abstract}
The rapid adoption of Generative AI, including LLM-based chatbots like ChatGPT, has highlighted the need for accessible ways to support public understanding and AI literacy. To address this need, we introduce a game-based, interactive approach in the form of a jigsaw puzzle whose completed image is a comic-based infographic illustrating the workings, capabilities, limitations, and societal implications of these technologies. Each comic sketch also functions as a standalone informational card, providing focused explanations of specific facets of AI use, design, and impact. The visual content was created in a live collaborative session with a professional illustrator and a multidisciplinary group of experts and non experts, combining structured knowledge with informal, exploratory reflections shared during the discussion. By integrating hands-on assembly, visual storytelling, and collaborative interaction, the puzzle provides an engaging and playful tool for exploring the mechanisms, perks, and perils of AI systems in informal learning contexts.
\end{abstract}


\begin{figure}[htbp]
\centerline{\includegraphics[width=\columnwidth]{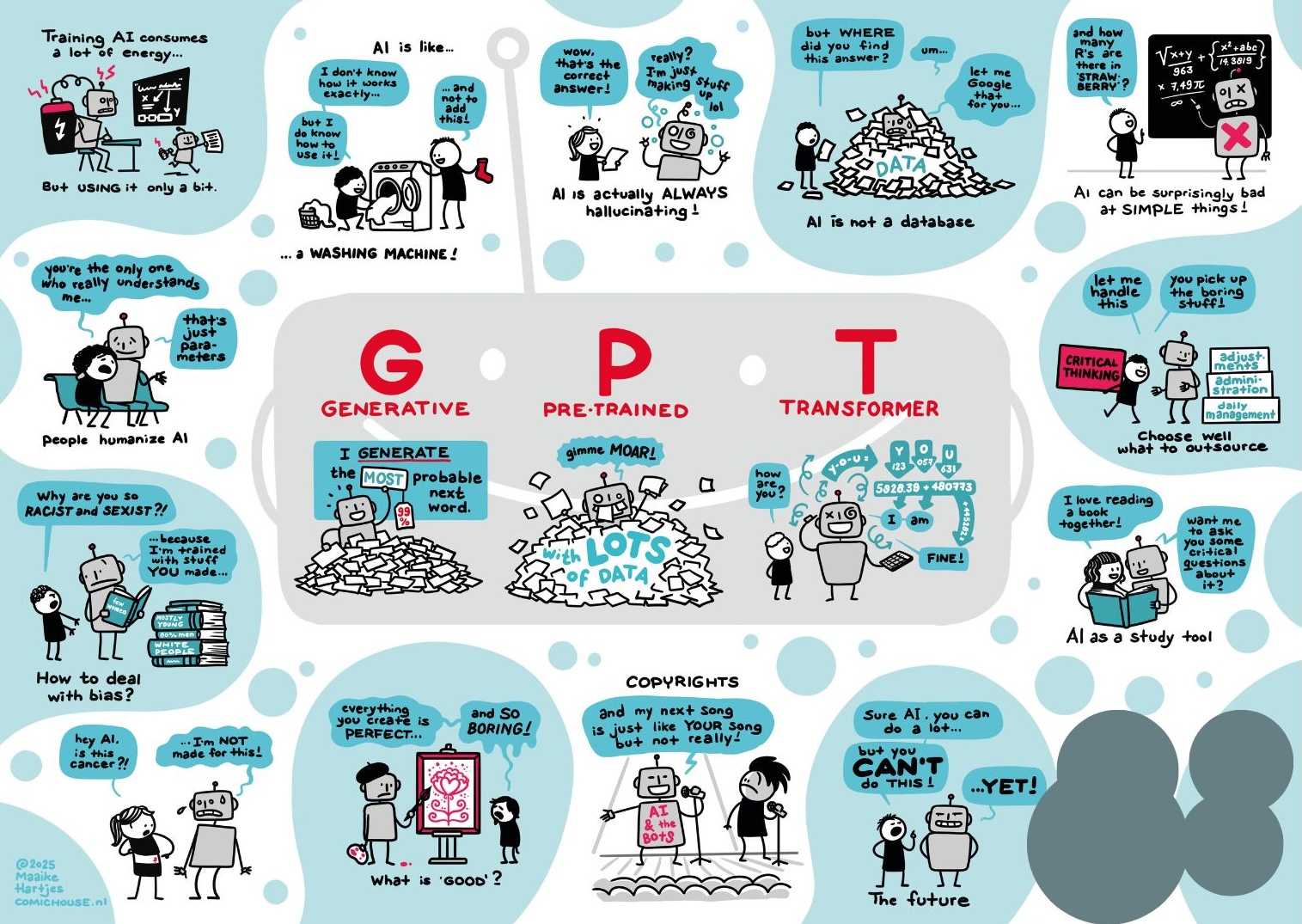}}
\caption{The complete jigsaw puzzle image\protect\footnotemark.}
\label{fig:puzzle_intero}
\end{figure}
\footnotetext{Patches in the bottom-right conceal institutional logos and funding acknowledgements to preserve author anonymity during review. The copyright notice in the bottom-left, attributed to the sketch artist, has been retained.}

\section{Introduction and Motivation}
The increasing pervasiveness of Generative Artificial Intelligence (GenAI), including Large Language Model (LLM)-based chatbots like ChatGPT \cite{openai2023}, across social, professional, and cultural contexts calls for greater attention to the development of AI literacy in the general public. This translates into the need for accessible and readily implementable frameworks and materials to support a broad understanding of these technologies, both within educational settings and among the general adult population, including teenagers and young 
adults \cite{renkema2025ai,hutler2025metacognitive}. This need is further reinforced by the speed and scale of the adoption of GenAI; these technologies have entered everyday life rapidly and unexpectedly, becoming deeply embedded in daily practices, often unknowingly for (and undisclosed to) the end user. By 2025, platforms such as ChatGPT had reached an unprecedented level of global diffusion, ranking among the most visited websites worldwide \cite{explodingtopics2025}. As a result, individuals are increasingly interacting with systems whose underlying mechanisms, limitations, and societal implications remain often poorly understood \cite{scantamburlo2024artificial,savoldi2025generative}.

Given the current landscape, fostering AI literacy beyond academic environments and across all segments of the population is no longer an intellectual aspiration, but rather a moral imperative, especially for practitioners of language technology. It is a necessary condition for the sustainable development of society during a moment of profound technological transition. The widespread adoption of GenAI systems is already transforming everyday practices, enabling new forms of personalised learning \cite{abbes2024generative, skjuve2024people}, increasing convenience in daily tasks, and affecting professional workflows \cite{noy2023experimental} by reshaping how individuals create content, access information, and carry out their work, while also raising questions about over-reliance, accountability, and the exercise of personal judgement \cite{brynjolfsson2025generative}.

In this context, non-formal approaches such as play offer a promising avenue to explore and reflect on the implications of GenAI systems. Drawing on Geertz’s \emph{Deep Play} concept \cite{geertz1972deep}, playful activities can reveal how social hierarchies and interactions are embedded in everyday practices, providing insight into how AI adoption may affect different communities and demographic groups. Learning through play fosters engagement, dialogue, and critical reflection, supporting collaborative problem-solving and thoughtful consideration of complex issues \cite{kanhadilok2014adult}. Play is increasingly recognised as a research and educational method, enabling experimentation, improvisation, and situated reflection \cite{van2018considering,leorke2026play}. When applied in practice, play can foster creativity, inspire new understandings, and encourage reflection on positionality, power, and responsibility, offering a flexible setting to explore how GenAI systems function and what their social impact is \cite{henricks2020play, frissen2025homo}.

Motivated by the need to support public understanding and awareness of GenAI systems, in particular conversational chatbots like ChatGPT, we sought to develop an outreach initiative to help non-expert users explore how these systems function, how they can be used, and the societal implications of their deployment. In this short paper, we present a project that combines the creative work of a professional illustrator with an innovative educational and playful approach to AI literacy. The outcome is a jigsaw puzzle centered on GPT-like language models, whose image consists of a comic-based infographic addressing key themes related to these systems, including their capabilities, limitations, risks, and opportunities. In the following sections, we describe the origin of the idea, the process through which it was developed, and the ways in which it will be tested, used, and further expanded in future dissemination efforts.

\section{How the idea of a jigsaw puzzle came about}

Jigsaw puzzles combine cognitive challenge, playful engagement, and collaborative interaction, making them highly effective for casual and self-reflective learning experiences \cite{garcia2016explorer}, with recognised educational benefits particularly for younger audiences \cite{williams2004jigsaw}. Studies in educational contexts have shown that puzzle-solving promotes problem-solving, abstract visualisation, mental modelling, and visual-spatial processing, while also fostering curiosity, persistence, and a sense of accomplishment \cite{rodenbaugh2014having}. Evidence from language education further highlights that collaborative puzzle-based interventions can enhance engagement, peer learning, and learning continuity, resulting in measurable improvements in knowledge acquisition compared to traditional methods \cite{layoc2025transforming}. Beyond cognitive benefits, puzzles also influence emotional and experiential dimensions: they can induce immersion, positive affect, and altered time perception across different age groups, enhancing motivation and reflective engagement \cite{garcia2016explorer, iwamoto2011alteration}. These combined cognitive, social, and emotional effects make jigsaw puzzles particularly suitable for exploring complex topics, such as the workings, capabilities, and societal implications of language-based AI systems, in a hands-on, engaging way.

In our project, the jigsaw puzzle, as shown in Figure~\ref{fig:puzzle_intero} presents a composite illustrated narrative that conveys the workings, capabilities, limitations, and societal implications of conversational agents like ChatGPT. The overall image consists of individual comic elements, each representing a self-contained informational unit. These elements can also be explored separately as cards, with in depth explanations on the reverse, which are provided alongside the puzzle to allow players to deepen their understanding of the visuals.

\begin{figure}[htbp] 
    \centering
    \begin{minipage}[b]{0.25\textwidth} 
        \centering
        \includegraphics[width=\textwidth]{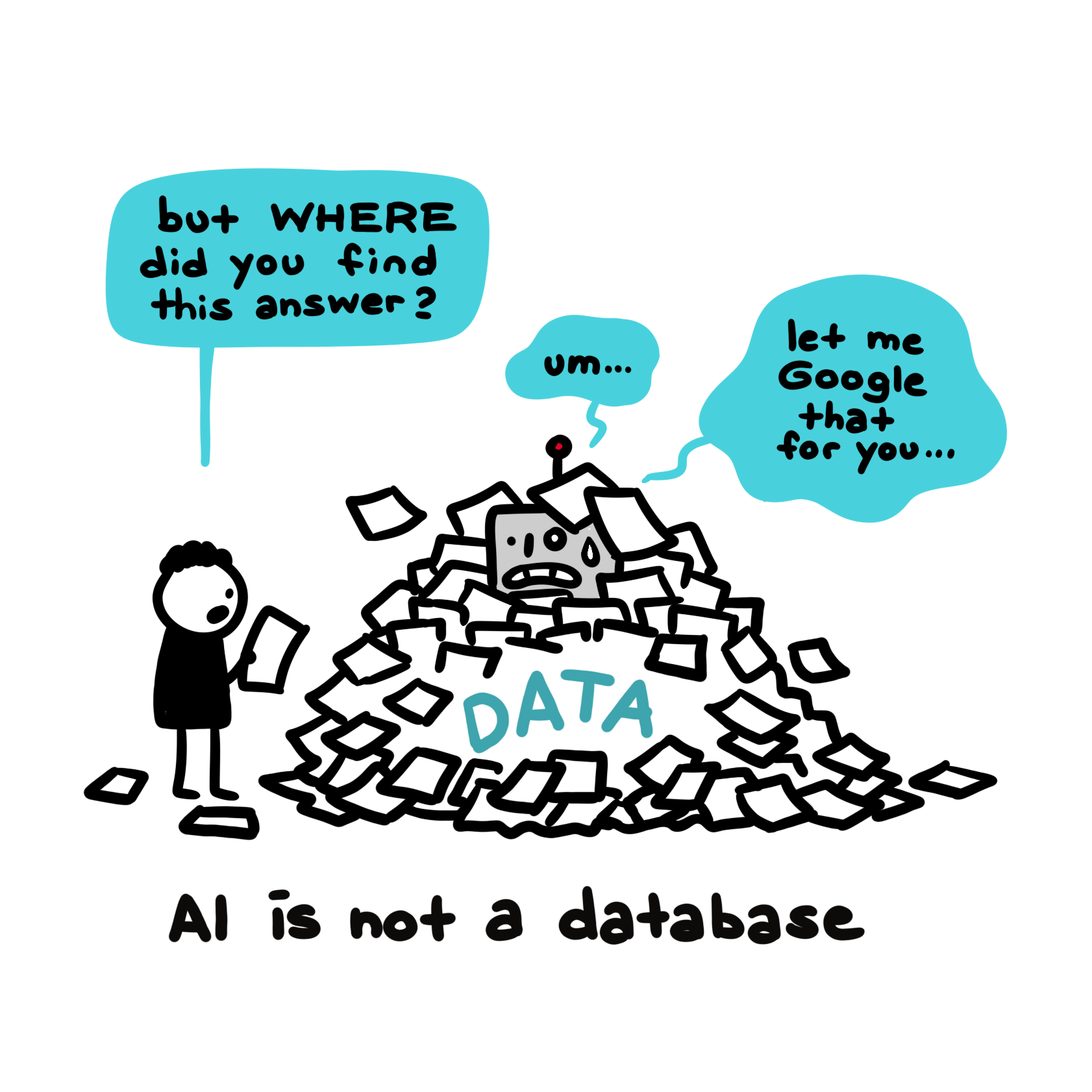}
    \end{minipage}%
    \begin{minipage}[b]{0.25\textwidth} 
        \centering
        \includegraphics[width=\textwidth]{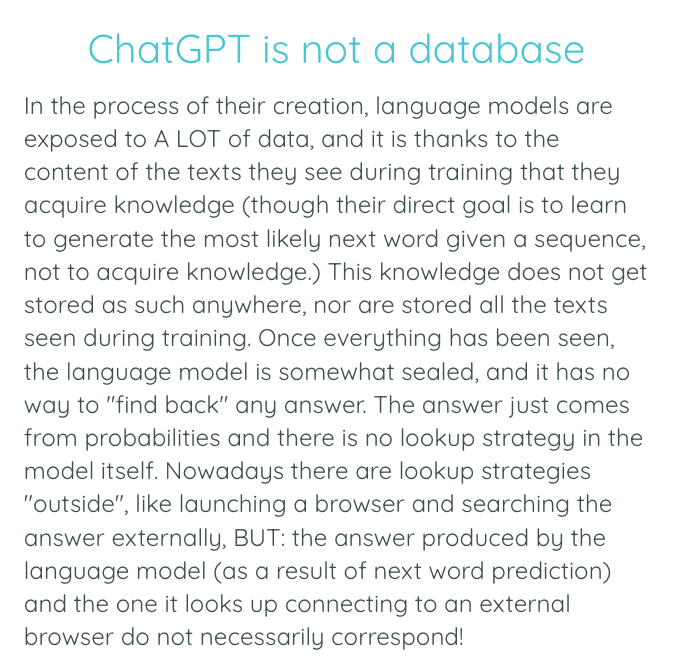}
    \end{minipage}%
    \caption{Front/back of a  card on the theme “ChatGPT is not a database”.}
    \label{fig:card_aidatabase}
\end{figure}

\begin{figure}[htbp]
    \centering
    \begin{minipage}[b]{0.25\textwidth}
        \centering
        \includegraphics[width=\textwidth]{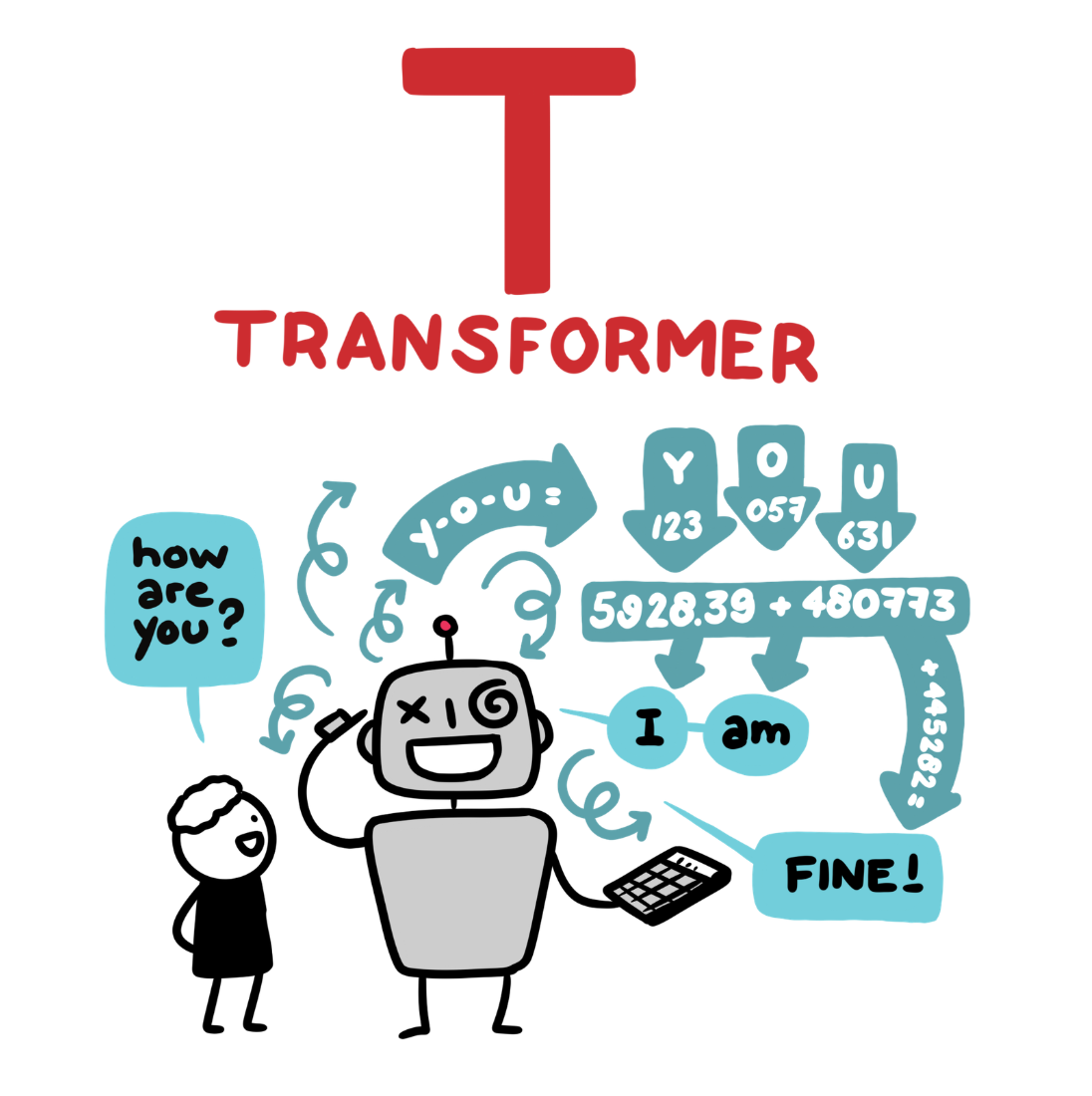}
    \end{minipage}%
    \begin{minipage}[b]{0.25\textwidth}
        \centering
        \includegraphics[width=\textwidth]{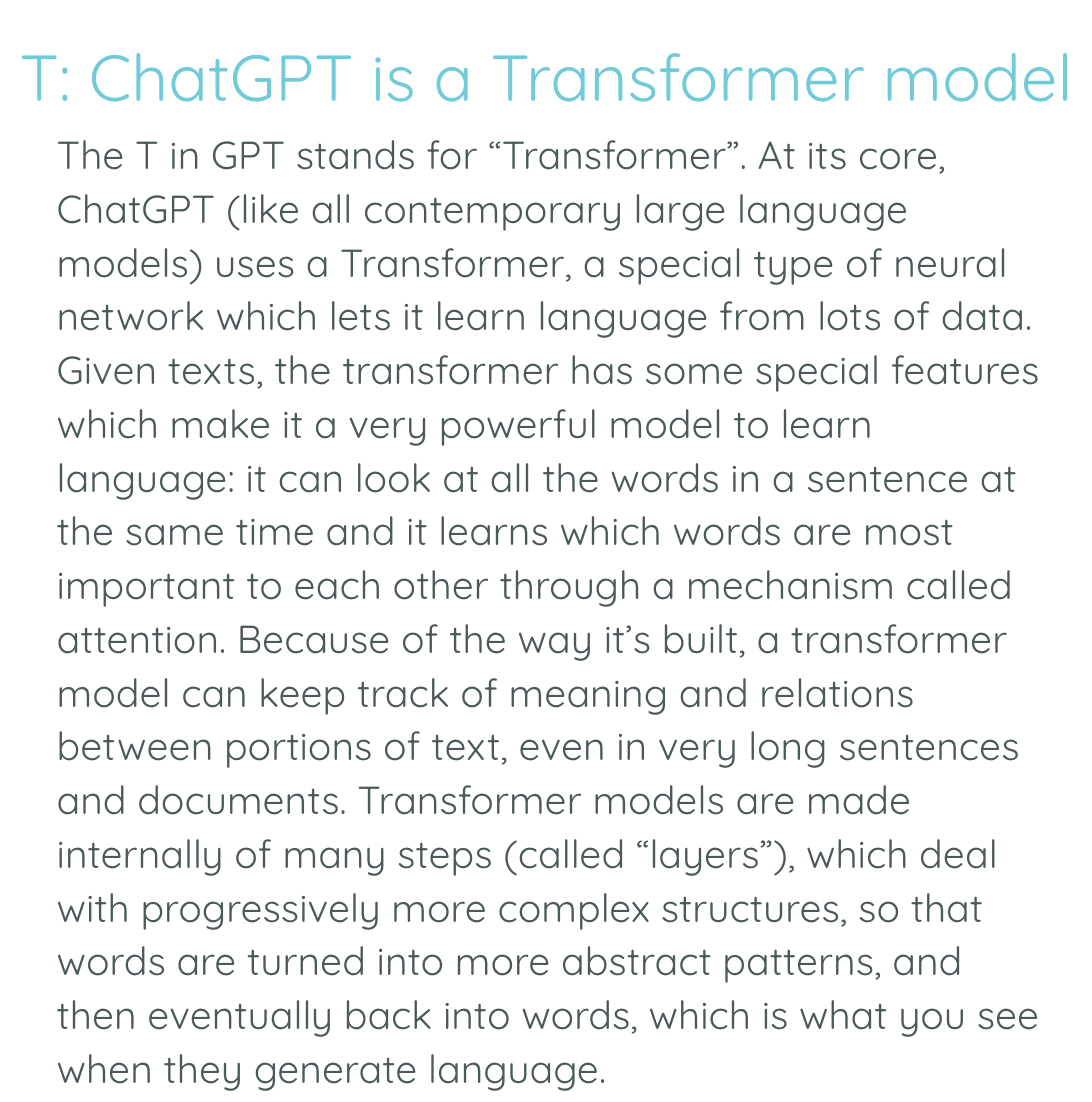}
    \end{minipage}%
    \caption{Front/back of a card introducing the \textit{Transformer} architecture.}
    \label{fig:card_transformer}
\end{figure}

Figures~\ref{fig:card_aidatabase} and \ref{fig:card_transformer} illustrate the front and back of two cards on \textit{ChatGPT not being a database} and the \textit{Transformer} architecture, showing how technical concepts are conveyed through visual metaphors and simple explanations.
By assembling the puzzle players repeatedly revisit key notions of AI systems, critically reflect on the material, and discuss topics such as biases in LMs, hallucinations, or copyright issues. Bringing together narrative elements, interactive assembly, and collaborative exchange, the activity encourages engagement and helps participants grasp and retain complex concepts.

\subsection{Co-Creating the Puzzle Visuals: Live Collaborative Session}

The visual content of the puzzle was developed in a dedicated live session, designed to translate multifaceted notions and thoughts about conversational agents and GenAI into comics-based narratives. During this session, participants---both experts and non-experts of AI---discussed a range of relevant topics, while a professional illustrator transformed the conversations into sketches in real time. A  list of topics had been precompiled and shared with all participants so that key concepts would not be missed in the discussion, but there was plenty of space for improvisation and for introducing new topics spontaneously during the live session itself.

\subsubsection{Illustrator Profile and Participation}
Maaike Hartjes\footnote{\url{https://maaikehartjes.nl/}} is a professional sketch artist known for distilling complex subject matter into sharp and humorous visual narratives. She listened to the dialogue in real time and translated emerging topics and insights into comic-style sketches in real time, which later formed the visual core of the puzzle image.

\subsubsection{Session Participants and Methodology}
The session brought together a multidisciplinary group composed of AI experts, such as computational linguists, and professionals from other fields, including digital communication, journalism, and law. Experts guided discussions on key topics related to AI, such as data sources, language diversity, bias, explainability, reasoning abilities, societal impact, and applications in education and professional contexts. Simultaneously, participants with expertise in other domains contributed personal reflections and experiences, enriching the conversation with diverse perspectives, critical viewpoints, and informal observations. Over approximately four hours, the artist produced around fifty sketches capturing both planned topics and unstructured reflections shared during the discussion. The sketches were then reviewed collaboratively in the course of the following weeks; members of the live session provided feedback and suggestions regarding the generated sketches and their visual content, while the artist made the final decisions on which elements would be incorporated into the final puzzle picture, as well as their placement, composition, and visual coherence. This approach combined structured knowledge with participants’ critical reflections and personal insights and style, with the artist's curatorial decisions ensuring a product that is at the same time informative, visually compelling, and unique. Of the original 50+ sketches produced during the live session, 31 were eventually developed into single cards, and 16 of them were used to compose the puzzle's image.

\subsubsection{Outcome}
The resulting final jigsaw puzzle provides an educational experience for those who explore it. As players piece together the puzzle, they repeatedly encounter key concepts about conversational agents. Many of the drawings feature subtle references and layered meanings, which often require discussion and exchange with other group members to be fully interpreted. Each sketch can also be explored individually as a standalone card accompanying the puzzle, where  detailed explanations of the physical cards can help clarify complex concepts. In addition to their physical format, the cards and accompanying explanations are accessible via the project’s website.\footnote{Link withheld to preserve anonymity.}, where the same explanations are reported, plus additional links for a more in-depth exploration of the topics, both with scientific and pop-culture materials.
This combination of collaborative interpretation and on-demand reference supports a deeper understanding of the technical mechanisms, potential benefits, and risks of GenAI systems, making the learning experience richer and more accessible, particularly for participants without prior expertise. The dynamic nature of the online materials allows for regular updates, well suited to the fast-paced nature of the field.

\begin{figure}[htb]
\includegraphics[width=\columnwidth]{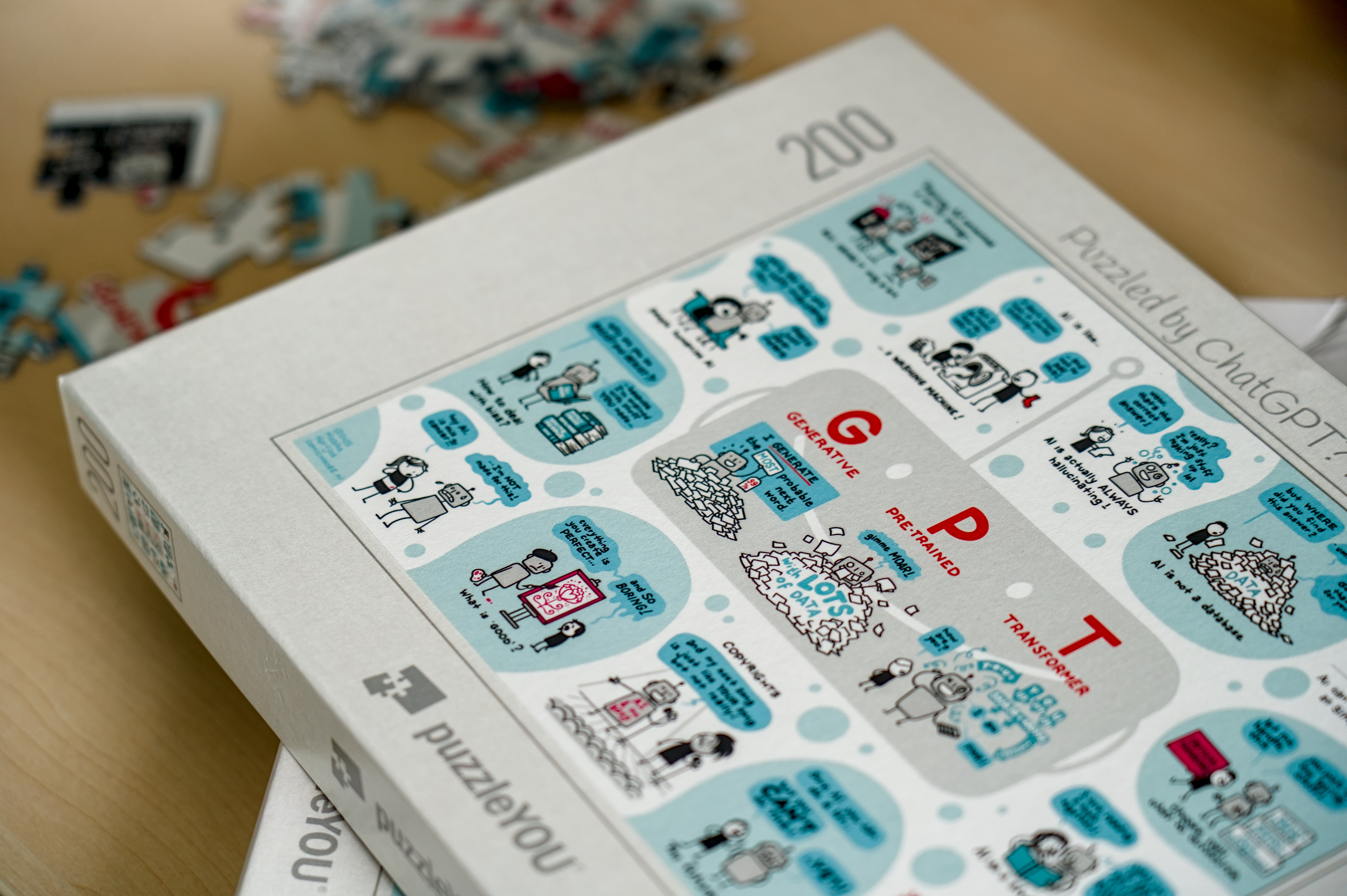}
\caption{Puzzle boxes and some pieces in the background.\label{fig:box}}
\end{figure}

The puzzle was printed in varying piece sizes: 200, 500, 1000, and 1000XXL, with the latter yielding a final object which is substantially larger than the other sizes, something that can be particularly appealing to some players. The pieces of the 200 version are larger which may make them easier to view and manipulate by, e.g., older people. Figure~\ref{fig:box} offers an anonymity-preserving peek of the boxes.

\subsection{First test with the public: The European Researchers' Night}


The European Researchers' Night is an annual European event sponsored by the European Commission, bringing scientific research closer to the general public through interactive activities. The session described in this paper took place in late 2025 as part of a local edition of the event, hosted in a multi-level cultural venue transformed into a dynamic exhibition space, where visitors could explore various scientific domains through talks, demonstrations, and hands-on activities.

\begin{figure}[htbp]
    \centering
    \begin{minipage}[b]{0.18\textwidth}
        \centering
        \includegraphics[width=\textwidth]{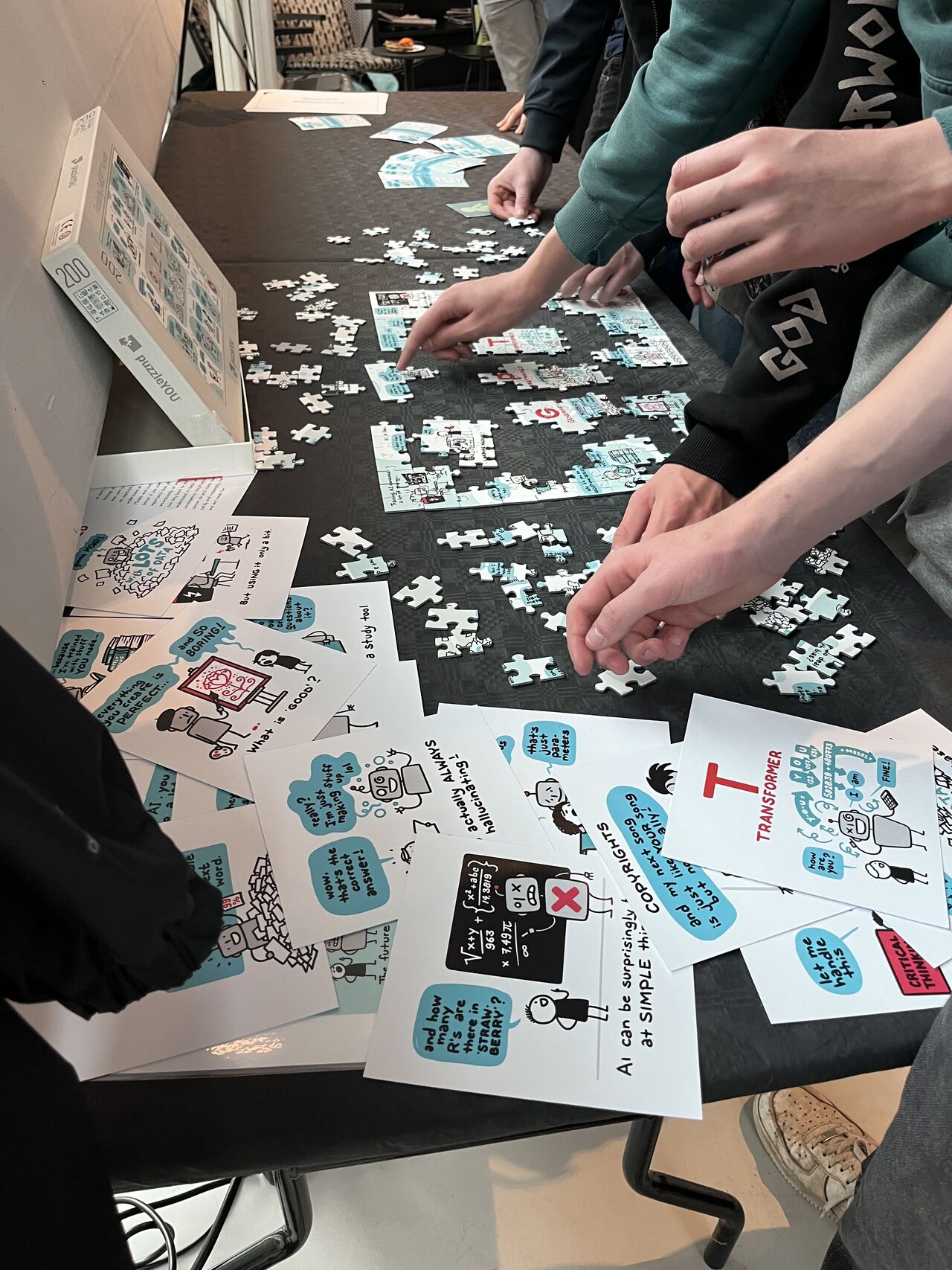}
    \end{minipage}%
    \hspace{0.05\textwidth}%
    \begin{minipage}[b]{0.18\textwidth}
        \centering
    \includegraphics[width=\textwidth]{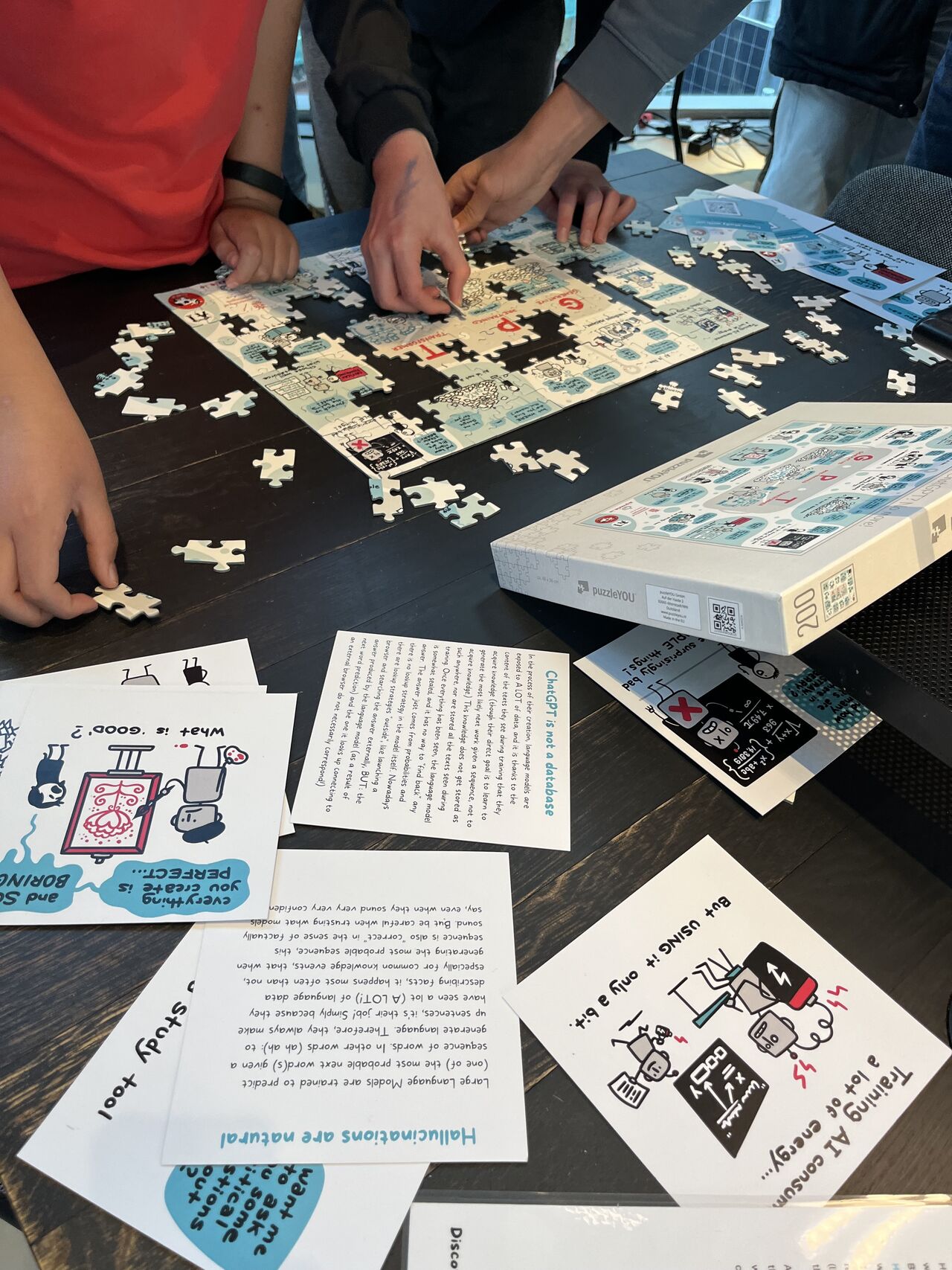}
    \end{minipage}%
    \caption{Puzzle stations during the European Researchers' Night.}
    \label{fig:rn}
\end{figure}

On this occasion, we presented the jigsaw puzzle to the public for the first time, situating it within an open and participatory environment. Throughout the event, two puzzle stations (200 pieces each\footnote{The larger size of the pieces and their limited number allow for easier completion of the puzzle in the short time each group visited our stations.}) were available to visitors, as shown in~\ref{fig:rn}. The first part of the day was dedicated to classes of high school students, while the second part was open to the general public, where groups of friends and families interacted with the puzzle, explored its contents, and took part in assembling it. After a brief introduction to the project and with guidance from the research team, participants used the accompanying cards to support their understanding of the puzzle. This setup facilitated discussions on LMs and their broader implications, with the visual elements of the puzzle serving as effective entry points for exploring technical concepts as well as broader social and ethical dimensions. A recurring observation made by the research team was the important role of the accompanying cards helping to unpack the complexities of the illustrated content. While the puzzle itself stimulated curiosity and interaction, the textual explanations on the back of the cards provided guidance on subtler points, linking the illustrations to the technical mechanisms and broader implications.

It is important to note that this initial deployment was exploratory in nature. No structured data collection was conducted, as the primary goal was to observe how the puzzle was received by a broad and diverse audience. The insights gathered are therefore qualitative and informal, focusing on general impressions and interaction patterns rather than measurable learning outcomes. Overall, the reception was highly positive. Several visitors expressed strong interest in the project, with approximately ten individuals requesting to be contacted in order to purchase a copy. This initial feedback suggests that the format is both appealing and accessible to a wide audience. Following this first informal test, the need emerged to more systematically investigate the educational potential of the puzzle, including its impact on users’ understanding and perception of GenAI systems, as well as the quality of the interaction experience. This motivated the design of a more structured and large-scale evaluation phase (see Section~\ref{sec:evaluation}).

\section{User Experience and Learning Outcomes}
\label{sec:evaluation}

We are currently conducting an evaluation of the puzzle experience to investigate participants' engagement, enjoyment, and perceived learning. This will provide insights into how game-based, collaborative activities can support AI literacy and critical reflection on GenAI systems. Findings will expand into a more in-depth publication focusing on learning outcomes and pedagogical potential in informal learning contexts.

\paragraph{Participants and Setup}
Participants are organised into small groups of three to ten members (friends, family, or colleagues), free to choose their preferred puzzle size, location, and time. Groups receive the box with the exploratory cards and assemble the puzzle collectively, encouraged to think critically and share observations throughout the process.

\paragraph{Data Collection}


After completing the puzzle, each participant fills out an individual online questionnaire covering demographics, prior AI familiarity, and digital skills. Beyond rating their overall experience, participants report on their perceived understanding of GenAI concepts and their awareness of ethical and societal implications of these technologies following the activity.

\medskip

The evaluation aims to:

\begin{itemize}
\item Assess participants’ engagement and enjoyment while assembling the puzzle.
\item Explore whether the puzzle enhances understanding of GenAI and LLM-based systems.
\item Investigate the development of critical awareness regarding AI-related risks and the potential consequences of AI technologies for society.
\item Examine how collaborative discussion and the interactive nature of the puzzle contribute to learning.
\item Collect feedback on the clarity, accessibility, and instructional value of the puzzle and accompanying cards.
\end{itemize}

By combining quantitative ratings with open-ended answers, this ongoing evaluation will provide  insights into the puzzle’s potential to support AI literacy in informal learning contexts.

\section{Discussion}

This work highlights the potential of playful, collaborative activities to foster AI literacy and critical engagement with LLM-based conversational agents such as ChatGPT. We believe in the power of play for its communal dimension and the cascading cognitive and reflective effects it generates. Through this paper, we aim to share our initial development efforts, inviting researchers and enthusiasts to explore similar approaches as a line of outreach complementing fundamental research. Such initiatives are essential for facilitating informed dialogue with the broader public on transformative technologies increasingly embedded in daily life.



\bibliographystyle{IEEEtran}
\bibliography{references}

\end{document}